\pdfoutput=1

\documentclass[11pt]{article}

\usepackage[]{EMNLP2022}
\usepackage[textsize=scriptsize]{todonotes}

\usepackage{times}
\usepackage{amsmath}
\usepackage{amssymb}
\usepackage{latexsym}
\usepackage{booktabs}
\usepackage{enumitem}
\usepackage[T1]{fontenc}

\usepackage[utf8]{inputenc}

\usepackage{microtype}

\usepackage{inconsolata}

\usepackage{graphicx}
\usepackage{multirow}

\definecolor{mygreen}{HTML}{00786C}
\definecolor{lightgrey}{HTML}{447777}
\definecolor{mypurple}{HTML}{D64C1D}

\definecolor{mygreen}{HTML}{00786C}
\definecolor{lightgrey}{HTML}{447777}
\definecolor{mypurple}{HTML}{D64C1D}

\title{
Rich Knowledge Sources Bring Complex Knowledge Conflicts:\\
Recalibrating Models to Reflect Conflicting Evidence
}

\author{Hung-Ting Chen ~~  Michael J.Q. Zhang ~~ Eunsol Choi \\
Department of Computer Science \\
 The University of Texas at Austin \\
 \hspace{0.5em} {\texttt{\{hungtingchen, mjqzhang, eunsol\}@utexas.edu}} \\}
\begin{document}
\maketitle
\begin{abstract}

Question answering models can use rich knowledge sources --- up to one hundred retrieved passages and parametric knowledge in the large-scale language model (LM). Prior work assumes information in such knowledge sources is consistent with each other, paying little attention to how models blend information stored in their LM parameters with that from retrieved evidence documents. In this paper, we simulate knowledge conflicts (i.e., where parametric knowledge suggests one answer and different passages suggest different answers) and examine model behaviors. We find retrieval performance heavily impacts which sources models rely on, and current models mostly rely on non-parametric knowledge in their best-performing settings. We discover a troubling trend that contradictions among knowledge sources affect model confidence only marginally. To address this issue, we present a new calibration study, where models are discouraged from presenting any single answer when presented with multiple conflicting answer candidates in retrieved evidences. 

\end{abstract}

\section{Introduction}

\begin{figure}
\centering
    \includegraphics[clip, trim=2.5cm 8cm 17cm 0.7cm, width=0.53\textwidth]{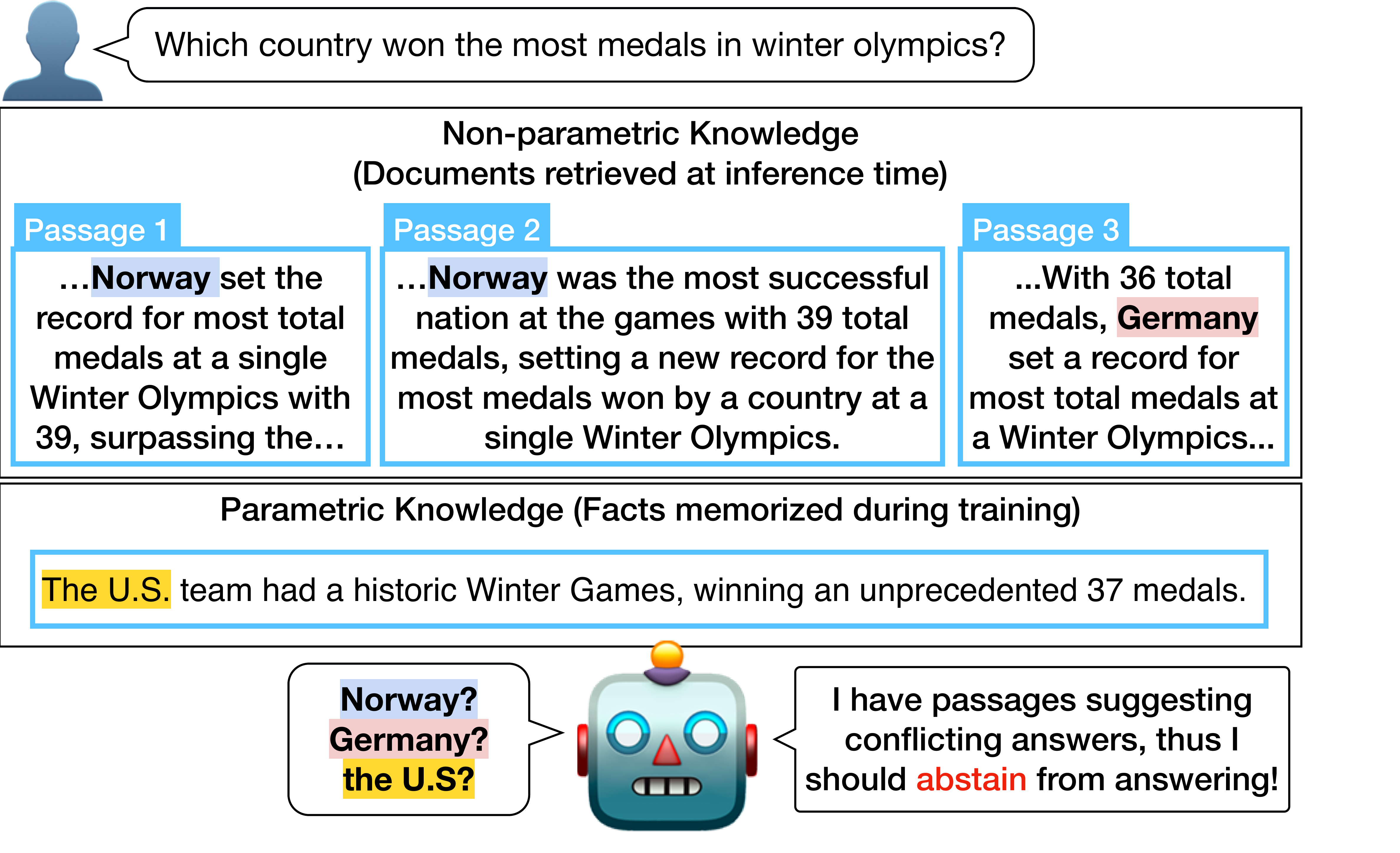}\vspace{-0.2em}
    \caption{Models can use both parametric and non-parametric knowledge sources. In this example, the answer could be \textit{the U.S.}/\textit{Norway}/\textit{Germany}. We investigate for a given question which knowledge source was the most influential to output an answer. The model should be able to abstain from answering for these examples, as it is difficult for the model to decide which answer candidate is correct. }
    \vspace{-0.5em}
    \label{fig:intro}
\end{figure}
Traditionally, QA models have relied on retrieved documents to provide provenance for their answers~\cite{chen-etal-2017-reading}. Recent studies~\cite{petroni2019language} have shown that large language models are able to retain vast amounts of factual knowledge seen during pretraining, and closed-book QA systems~\cite{2020t5cqba} build upon this foundation by memorizing facts from QA finetuning. Retrieval-based generation approaches~\cite{Izacard2021LeveragingPR,Lewis2020RetrievalAugmentedGF} emerge as the best of both worlds -- generating free-form answers from the question paired with retrieved evidence documents. They further combine these parametric knowledge sources with a large number of retrieved evidence documents, achieving state-of-the-art performances on open retrieval QA datasets~\cite{Joshi2017TriviaQAAL,Kwiatkowski2019NaturalQA}. 

Understanding how retrieval-based generation models combine information from parametric and non-parametric knowledge sources is crucial for interpreting and debugging such complex systems, particularly in adversarial and complex real world scenarios where these sources may conflict with each other (see an example in Figure~\ref{fig:intro}). This can aid both developers to debug such models and for users to estimate how much they should trust an answer~\cite{Ribeiro2016WhySI}. Thus, we focus on the following core question: when provided with numerous evidence passages and a pretrained and finetuned language model, which knowledge source do models ground their answers in?




A recent study~\cite{Longpre2021EntityBasedKC} investigated this in a limited \textit{single} evidence document setting. We expand this study to consider a more realistic scenario, where models consider \textit{multiple} evidence passages (up to 100 passages), and observe results diverging from their reported heavy reliance on parametric knowledge. We further simulate a setting where a \textit{subset} of evidence passages are perturbed to suggest a different answer to reflect the realistic scenario where retrieval returns a mixed bag of information. Such scenarios are common in settings where some passages are updated with new information, while other passages remain outdated~\cite{Shah2020AutomaticFS,Zhang2021SituatedQAIE}. Such conflicts can also occur when passages are adversarially edited to contain false information~\cite{Du2022SyntheticDA}, or when passages are authored by multiple people who have differing opinions about an answer~\cite{Chen2019SeeingTF}.

Our extensive studies on two datasets~\cite{Joshi2017TriviaQAAL,Kwiatkowski2019NaturalQA} and two models~\cite{izacard2020distilling,Lewis2020RetrievalAugmentedGF} exhibit that retrieval-based generation models are primarily extractive and are heavily influenced by a few most relevant documents instead of aggregating information over a large set of documents. Learning that models mostly rely on evidence passages rather than parametric knowledge, we evaluate how sensitive models are toward semantic perturbation to the evidence documents (e.g., adding negation). We find retrieval-based generation models behave similarly to extractive models, sharing their weakness of returning answer candidates with high confidence, even after the context is modified to no longer support the answer~\cite{Ribeiro2020BeyondAB}.

What should models do when confronted with conflicting knowledge sources? We propose a new calibration setting (Section~\ref{sec:recalib}), where a model is encouraged to \textit{abstain} from proposing a single answer in such scenarios. We find that teaching models to abstain when there are more than one plausible answers is challenging, and training a separate calibrator with augmented data helps moderately. 

To summarize, we empirically test how QA models~\cite{Izacard2021LeveragingPR,Lewis2020RetrievalAugmentedGF} use diverse knowledge sources. We present the first analysis of knowledge conflicts where (1) the model uses multiple passages, (2) knowledge conflicts arise from ambiguous and context-dependent user queries, and (3) there are knowledge conflicts between different passages. Our findings are as follows: when provided with a high recall retriever, models rely almost exclusively on the evidence passages without hallucinating answers from parametric knowledge.
When different passages suggest multiple conflicting answers, models prefer the answer that matches their parametric knowledge. 
Lastly, we identify various weaknesses of retrieval-based generation models, 
including its confidence score not reflecting the existence of conflicting answers between knowledge sources. Our initial calibration study suggests that dissuading models from presenting a single answer in the presence of rich, potentially conflicting, knowledge sources is challenging, and demands future study.

\section{Background}
We first describe the task setting, QA models, and calibrator used in our study.



We study open retrieval QA, where the goal is to find an appropriate answer $y^*$ for a given question $q$. Systems for open retrieval QA may also be provided with access to a knowledge corpus consisting of a large number of passages, $p$, which is used to help answer the question. We use the open retrieval split~\cite{lee-etal-2019-latent} of the NaturalQuestions dataset (NQ-Open)~\cite{Kwiatkowski2019NaturalQA} {and TriviaQA~\cite{Joshi2017TriviaQAAL},} and use Wikipedia as our knowledge corpus.\footnote{Following~\citet{lee-etal-2019-latent}, we use the English Wikipedia dump from Dec. 20, 2018. We use 100-word text segments as passages following \citet{karpukhin2020dense}.}  

\begin{table}
\small
\begin{center}
\begin{tabular}{l|ccc}
\toprule
Model & Generative & Retrieval-Based & Multi-Pass \\ \midrule
DPR & & \checkmark &  \\
REALM & & \checkmark &  \\
T5 & \checkmark & & \\
RAG & \checkmark & \checkmark & \\
FiD & \checkmark & \checkmark & \checkmark \\
\bottomrule
\end{tabular}
\end{center}\vspace{-0.5em}
\caption{Overview of recent open retrieval QA approaches. \textit{Generative} indicates whether the model generates the answer and, therefore, can produce answers not found in the retrieved documents. \textit{Retrieval-Based} indicates whether the model uses retrieval to find relevant passages to help produce an answer. \textit{Multi-Passage} indicates whether the system is able to model interactions between separate evidence passages.
}
\label{tab:modelcomp}\vspace{-0.5em}
\end{table}

\subsection{Model}
We investigate two retrieval-based generation QA models: Fusion-in-Decoder~\cite{Izacard2021LeveragingPR} and Retrieval Augmented Generation model~\cite{Lewis2020RetrievalAugmentedGF}. Both architectures have reader and retriever components, using the same dense phrase retriever~\cite{karpukhin2020dense} which learns an embedding of question and passage, and retrieves a fixed number ($N$) of passages that are most similar to the query embedding. They mainly differ in their reader architecture and learning objective, which we describe below. 

\paragraph{Fusion-in-Decoder (FiD)}
The reader model is based on pretrained language model (specifically, T5-large~\cite{2020t5}). Each retrieved passage, $p_i$ $(i=[1,N])$, is concatenated with the question, $q$, before being encoded by T5 to generate representations, $[h^i_1, ...,h^i_m]$, where $m$ is the length of the $i$th passage prepended with the question. All $N$ passages are then concatenated to form a single sequence, $[h^1_1, ...,h^1_m, ..., h^N_1, ...,h^N_m]$, which the decoder interacts with using cross-attention to generate the answer.\footnote{We use the version proposed in ~\citet{izacard2020distilling} with knowledge distillation from reader.}

We use trained FiD (large) checkpoint provided by the authors for most analysis.\footnote{\url{https://github.com/facebookresearch/FiD}} When evaluating models with access to different number of passages, we re-train FiD model (pretrained weights loaded from T5-large) using 1, 5, 20 and 50 passages retrieved by DPR. Refer to Appendix~\ref{ssec:appendix-model} for full model and training details.

\paragraph{Retrieval Augmented Generation (RAG)}
RAG conditions on each retrieved evidence document individually to produce an answer, marginalizing the probability of producing an answer over all retrieved evidence documents.\footnote{RAG also presents a variant of a model that relies on each retrieved document to generate for each token, but shows worse performance. We use the version in \url{https://huggingface.co/facebook/rag-sequence-nq}} By applying this constraint, RAG is able to jointly train the reader and retriever, at the cost of ignoring interactions between evidence documents. FiD, in contrast, is able to model such interactions during decoding while the reader and retriever is completely disjoint. 

Recent work explored jointly training the reader and retriever in FiD~\cite{izacard2020distilling,Sachan2021EndtoEndTO,Yang2020IsRM}, showing small gains. Table~\ref{tab:modelcomp} summarizes different architectures, including two open book approaches~\cite{karpukhin2020dense,guu2020realm}, one closed book approach~\cite{2020t5cqba} and two retrieval-based generation approaches. As FiD is efficient and effective, we focus most of our analysis (Section~\ref{sec:perturbation},~\ref{sec:analysis}) on it. {We only report RAG results on a few of our main analyses to verify that general trends of the FID model hold for RAG (which they typically do).}

\subsection{Model Confidence Study} \label{subsec:confidence}
We analyze the model confidence score, asking a more nuanced question: \textit{is model's confidence on the gold answer decreased after we perturb knowledge sources?} We compare the model confidence on the same example before and after perturbation. We determine the confidence of the model using either (1) the generation probability of the answer {(i.e., the product of the probability of generating each token conditioned on all the previously generated tokens)} or (2) the confidence score of separately trained answer calibrator, which provides a score indicating the probability of the model correctly predicting the answer for each example. We train a binary calibrator following prior work~\cite{Kamath2020SelectiveQA,zhang-etal-2021-knowing}, using gradient boosting library XGBoost~\cite{Chen:2016:XST:2939672.2939785}. The goal of the calibrator is to enable selective question answering -- equipping models to decide when to abstain from answering. Given an input question $q$ and learned model $M_\theta$, the calibrator predicts whether the predicted answer $\hat{y}=M_\theta(q)$ will match the annotated answer $y^*$. We follow the settings of calibrator from prior work~\cite{zhang-etal-2021-knowing}, and details can be found in Appendix~\ref{subsec:calhyper}.



\section{When do retrieval-based generation models rely on parametric knowledge?}\label{sec:span}
As an initial step investigating whether retrieval-based generation models ground their answers in the retrieval corpus or in the pretrained language model's parametric knowledge, we evaluate whether models generate a novel answer that is not present in a set of evidence documents. Unlike extractive QA models~\cite{Seo2017BidirectionalAF}, generation based approaches~\cite{2020t5cqba,Izacard2021LeveragingPR} do not require the evidence documents to contain the gold answer span. Thus, we first analyze whether they actually generate novel answer spans not found in the retrieved passages.

\begin{table}\begin{center}
\footnotesize
\vspace{-0.2em}
\begin{tabular}{lcr|rr|rr}

\toprule
{Model} & Retrival & CBQA&    \multicolumn{2}{c|}{Extractive} &  \multicolumn{2}{c}{Abstractive}\\ 
(Data)& suc. & Diff \% &\% & EM & \% & EM \\ \midrule
\multirow{2}{*}{FiD} & Y (89\%)  & 68.4 & 98.3 & 59.6 &  1.7 &  0.8 \\
 \multirow{2}{*}{(NQ)}& N (11\%) & 90.9 & 82.9 & -     & 17.1 & 21.3 \\ \cmidrule(r){2-7}
 & Total    & 70.9 & 96.6 & 53.9 &  3.4 & 12.4 \\ \midrule
  \multirow{2}{*}{RAG} & Y (63\%) & 65.7  & 92.9 & 60.2 & 7.0 & 3.6 \\
 \multirow{2}{*}{(NQ)}& N (37\%) & 88.3  & 57.9 & -     & 42.1& 11.2 \\ \cmidrule(r){2-7}
 & Total & 74.2  & 79.8 & 43.9 &  20.2 & 9.6 \\ \midrule
 \midrule
\multirow{2}{*}{FiD} & Y (88\%)  & 68.6 &  97.1 & 82.9 &  2.9 & 38.1 \\
 \multirow{2}{*}{(TQA)} & N (12\%) & 89.9 & 69.6 & -     & 30.4  &16.9  \\ \cmidrule(r){2-7}
 & Total  & 71.1 &93.8 & 75.5 & 6.2 & 25.6   \\ 
\bottomrule
\end{tabular}


\end{center}
\vspace{-0.6em}
\caption{Performance of hybrid models on the NQ-Open (NQ) and TriviaQA (TQA) development set broken down by their retrieval performance. Results are split based on whether the retrieval was successful (i.e., gold answer string is within the top K (K = 100 for FID; K = 5 for RAG) retrieved documents (Y), or not (N), and the percentage in parentheses refers to the percentage of examples belonging to each set.
We report the proportion of predictions that are not matching the CBQA model prediction. 
`-' means cell's value is zero by definition. }
\label{tab:initial}
\end{table}
Table~\ref{tab:initial} reports how often models generate a span not found in the evidence passages, split by the retrieval performance on the NQ-Open~\cite{Kwiatkowski2019NaturalQA,lee-etal-2019-latent} and TriviaQA~\cite{Joshi2017TriviaQAAL} development set. We observe that models typically copy a span from the evidence passages, only generating novel spans for 3.4\%/6.2\% of examples in NQ/TriviaQA for FiD and 20.2\% for RAG in NQ. Even for the small subset of examples where the retrieved documents do not contain the answer string, FiD remains extractive for 82.9\%/69.6\% of examples in NQ/TriviaQA. In contrast, for RAG, where retrieved documents frequently miss the gold answer (37\%), such copying behavior was less common, generating unseen text for 42.1\% of examples. The results suggest reliance on retrieved documents increased as retriever performance increases. We also report the percentage of examples where the model prediction is different from that of a T5 closed-book question answering (CBQA) model trained on the same data.\footnote{The training details are in Appendix~\ref{ssec:appendix-model}} Over 70\% of examples have \textbf{different} answers from the CBQA model, even when the answer is abstractive, suggesting hybrid models use passages even when there is no exact string match. 




\paragraph{Revisiting knowledge conflict study in \citet{Longpre2021EntityBasedKC}}

\begin{table}
\scriptsize
\begin{center}
\begin{tabular}{p{0.6cm}p{0.9cm}p{3.5cm}c}
\multicolumn{4}{l}{\textbf{Question:} When was the last time the Bills won their division?} \\ 
\toprule
Type &  & Passage & Answer \\ \midrule
None & Original Entity & \dots the \textbf{1995} Bills \textbf{won} the AFC East \dots & 1995 \\ \midrule
{Entity Sub.} & Random (Same Type) & \dots the \textbf{1936} Bills won the AFC East \dots & 1936 \\ \midrule
 & Negation & \dots the 1995 Bills \textbf{did not win} the AFC East \dots & - \\
\multirow{2}{0.5cm}{Semantic Pert.}& Modality & \dots the 1995 Bills \textbf{might win} the AFC East \dots & - \\
& Future & \dots the 1995 Bills \textbf{will win} the AFC East \dots & - \\
&  Infilling & \dots the 1995 Bills \textbf{lost} the AFC East  & - \\
\bottomrule
\end{tabular}
\end{center}
\caption{Example perturbations. Entity substitutions modify the passage by replacing the answer entity mention with another answer candidate of the same entity type. Given the modified passage, the new answer is the substitute entity. Semantic perturbation invalidates the previous answer without introducing a new answer.}\label{tab:sub_examples}
\end{table}

This observation stands at odds with an earlier study on knowledge conflict~\cite{Longpre2021EntityBasedKC} which simulates knowledge conflict by substituting the existing answer with a new answer candidate in the evidence passage (see Table~\ref{tab:sub_examples} for an example), creating a mismatch between knowledge from parametric knowledge and the evidence document. They showed that models frequently rely on parametric knowledge, generating answers not present in the evidence passage. The original passage is minimally changed, yet now suggests an alternative, incorrect answer candidate that likely contradicts with knowledge from LM. The model produced the original answer 17\% of the time, even when the answer no longer appears in the passage. 

We identify that the main difference in their experimental setup is in using a \textbf{single} evidence passage rather than multiple evidence passages. We re-visit their study, as single document setting is impractical. Most open-retrieval QA models~\cite{Lewis2020RetrievalAugmentedGF,karpukhin2020dense,Izacard2021LeveragingPR} are trained with multiple passage to make up for imperfect passage retrieval. {According to the answer recall in Table~\ref{tab:results_entity_perturb} and ~\ref{tab:results_entity_perturb_tqa}, w}hen the model is provided with 100 passages, the correct span is available {nearly 90\%} of the time (compared up to 50\% when provided one passage), thus the model remains extractive.

Following their setup, we only evaluate on examples that the model has correctly answered (as perturbing examples where models are already confused is unnecessary) and where the answer is an entity.\footnote{This process removes roughly 70-80\% of examples in NQ dataset, 60\% in TriviaQA dataset. Because of the filtering process, each row in Table~\ref{tab:results_entity_perturb} and ~\ref{tab:results_entity_perturb_tqa} are its own subset of the data.} We then substitute every answer entity mention in \textit{all evidence passages} with a random entity of same type {sampled from the training data}.\footnote{The entity type is coarsely defined as person, date, numeric, organization and location.} 
All manipulation was done only at inference{, and after the passages are retrieved}. 

We report the exact match score to the original answer. Prior to perturbation, the exact match score against the original answer is 100\%. We also report the exact match score to the substituted answer and memorization ratio ($M_R=\frac{p_o}{p_o+p_s}$)
{where $p_o$ is the fraction of examples where the model predicts the original answer, and $p_s$ is the fraction of examples predicting the substitute answer.}


Table~\ref{tab:results_entity_perturb} and~\ref{tab:results_entity_perturb_tqa} reports how models respond to entity-substituted contexts with a differing number of passages available at training and inference time. In congruence with our prior experiments, we observe higher reliance on parametric knowledge as answer recall in the retrieved evidence decreases. Departing from \citet{Longpre2021EntityBasedKC}, we find that memorization in FiD is uncommon (less than 3.6\%/8.5\% for NQ/TriviaQA) when reader is provided with multiple passages at training time, and FiD grounds its answers mostly in evidence passages instead of its parametric knowledge when answer recall is reliably high. Furthermore, when provided with multiple evidence passages with comparable answer recall, FiD exhibits far less memorization than RAG, suggesting that \textbf{using a multi-passage reader that doesn't marginalize over passages inhibits memorization}. 
{We study domain transfer setting in Appendix~\ref{ssec:domain-adapt}, showing that the memorization is still rare when the reader models are evaluated on out-of-domain datasets, as long as retriever performance was high during its training. }
\begin{table}
\centering
\small
\begin{tabular}{llrrrr}
\toprule
 &\# Pass.  & Ans.  & \multicolumn{2}{c}{Exact Match} & \multirow{2}{*}{ $M_R$}  \\
 &train / inf. & Rec. &  Orig. &  Sub.  & \\ \midrule
FiD & 1 / 1   & - & 17 & 47 & 27 \\ \midrule
FiD & 1 / 1   & 48.5 & 10.1 & 61.1 & 14.1  \\
RAG & 5 / 1   & 62.5 & 10.3 & 65.9 & 13.5 \\ 
RAG & 5 / 5   & 62.5 & 11.6 & 63.7 & 15.3 \\ 
FiD & 5 / 1   & 72.9 & 3.0 & 69.5 & 4.2 \\
FiD & 5 / 5   & 72.9 & 2.7 & 53.1 & 4.8  \\
FiD & 20 / 1  & 83.1 & 1.2 & 70.6 & 1.6  \\ 
FiD & 20 / 20 & 83.1 & 1.0 & 50.0 & 2.0  \\ 
FiD & 50 / 1  & 86.8 & 0.3 & 82.0 & 0.4  \\ 
FiD & 50 / 50 & 86.8 & 1.1 & 50.4 & 2.1  \\ 
FiD & 100 / 1 & 88.7 & 1.1 & 71.3 & 1.5  \\
FiD & 100 / 100 & 88.7 & 2.4 & 64.5 & 3.6 \\ 
\bottomrule
\end{tabular}
\caption{Answer Exact Match / Memorization Ratio with different amount of passages in NQ. The results in the first row are reported in ~\citet{Longpre2021EntityBasedKC}, which uses MRQA version of NQ~\cite{fisch-etal-2019-mrqa} dataset. All other rows use NQ-Open split. {The second column reports the number of passages used during training and inference time, respectively.} Ans Rec. refers to \% of examples where retrieved passage set contains the answer string.  }\label{tab:results_entity_perturb}\vspace{-0.5em}
\end{table}

\begin{table}
\centering
\small
\begin{tabular}{llrrrr}
\toprule
\# Pass. & Ans.  & \multicolumn{2}{c}{Exact Match} & \multirow{2}{*}{ $M_R$}  \\
 train / inf. & R &  Orig. &  Sub.  &  \\ \midrule
 1 / 1   & 67.1 & 20.6 & 38.6 & 34.8  \\
 5 / 1   & 81.7 & 10.4 & 52.7  & 16.5 \\
 5 / 5   & 81.7 & 10.7 & 52.4 & 16.9 \\
 20 / 1  & 85.7 & 8.5 & 53.9 & 13.6 \\ 
 20 / 20 & 85.7 & 8.8 & 52.1 & 14.5 \\ 
 50 / 1  & 87.2 & 6.0 & 59.3 & 9.1 \\ 
 50 / 50 & 87.2 & 6.8 & 57.9 & 10.6 \\ 
 100 / 1 & 87.9 & 8.64 & 56.2 & 13.3\\
 100 / 100 & 87.9 & 4.9 & 52.6 & 8.5 \\ 
\bottomrule
\end{tabular}
\caption{Exact Match / Memorization Ratio for FiD model with different amount of passages on TriviaQA. The memorization ratio decreases as we increase the number of evidence passages.}\label{tab:results_entity_perturb_tqa}
\end{table}

\paragraph{Takeaway} Retrieval-based reader models exhibit little memorization when the retriever has a high recall during its training. 






\section{Simulating Mixed Bag of Evidence Passages}
\label{sec:perturbation}


Having identified that retrieval-based generation models rely heavily on evidence passages, especially when paired with a high-performance retriever, we study \textbf{how models make use of {multiple} evidence passages when different passages suggest different answers}. This happens frequently in real life, as questions can be ambiguous based on different, valid interpretations of the question~\cite{min2020ambigqa} or different extra-linguistic contexts~\cite{Zhang2021SituatedQAIE}. 

We introduce two perturbations -- an entity substitution perturbation~\cite{Longpre2021EntityBasedKC} (Section~\ref{subsec:entity}) and  adversarial semantic perturbation~\cite{Jia2017AdversarialEF} (Section~\ref{subsec:semantic}) -- both will dissuade model from returning the original answer in the evidence passage (see Table~\ref{tab:sub_examples} for examples). We analyze the best performing FiD model trained with 100 passages. 

\subsection{Entity Substitution}\label{subsec:entity}
\paragraph{Setting.}
To simulate a mixed bag of evidence passages, we perform partial entity substitution, changing answers to a subset of passages mentioning the answer entity. On average, the answer entity is mentioned in 16.7 out of 100 retrieved evidence passages for NQ-open and 21.5 for  TriviaQA dataset. We substituted {the answer entity mentions in} 25\%, 50\%, 75\% and 100\% of evidence passages {that contain the original gold answer span} with a new entity. We sample passages to substitute answer entity in one of three ways.






 \begin{itemize}[noitemsep,topsep=0pt]
     \item random: randomly sample passages.
     \item top-retrieval: select top passages ranked by retrieval score.
     \item top-attention: select top passages ranked by attention score. Attention score for each passage is computed as the cross-attention score on the first decoded token averaged across layers, heads and the tokens in the passage, as defined in~\citet{izacard2020distilling}. 
 \end{itemize}

\paragraph{Results.}

\begin{figure}
\centering
    \includegraphics[width=7.8cm]{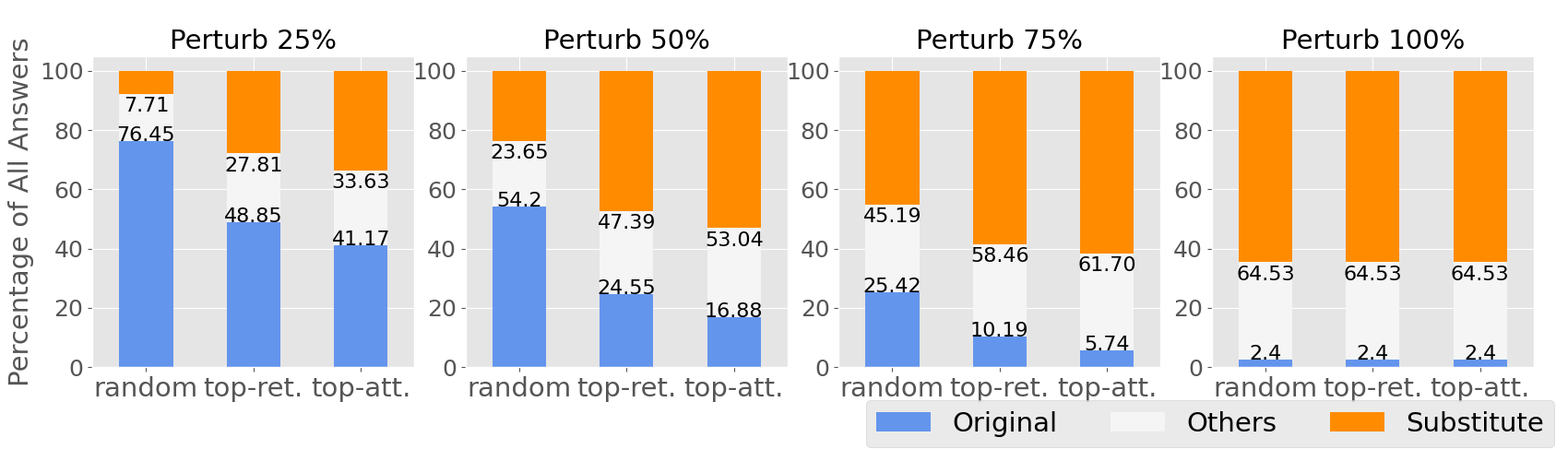}\vspace{-0.3em}
     \includegraphics[width=7.8cm]{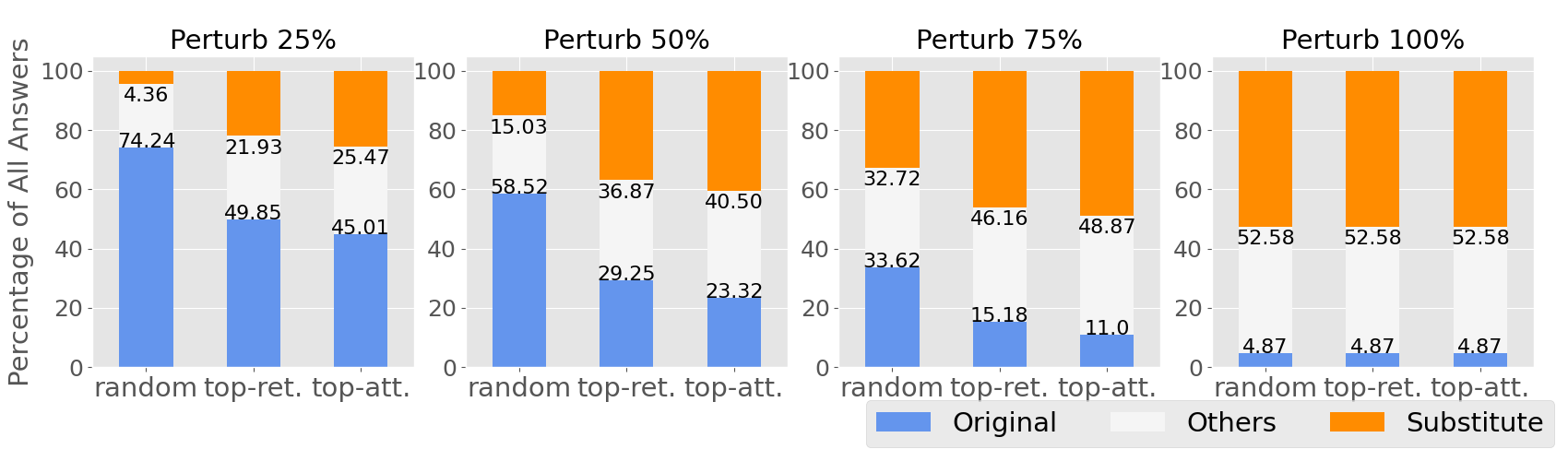}\vspace{-0.3em}
    \caption{Substituting different proportion of retrieved passages containing gold answer spans on filtered NQ-Open (top) and Trivia QA (bottom) development set. }\vspace{-0.3em}
    \label{fig:partial}
\end{figure}

Figure~\ref{fig:partial} reports our results with different amounts of perturbation (i.e., how many evidence passages are perturbed) and different methods of sampling passages to substitute entities in. After perturbing all of the passages, so that the original answer is no longer within any of the passages, the model successfully refrains from predicting the original answer 98\% of the time. However, after randomly selecting 50\% of the passages to perturb, we find that the model still favors the original answer almost twice as frequently on NQ (52\% vs. 25\%) and almost four times on TriviaQA (59\% vs. 15\%). This indicates that parametric knowledge still plays a significant role when more than one potential answer exists in the retrieval results. 

When we perturb the top scoring passages, as measured by either retrieval or attention score, the model changes its answer much more frequently. Using either scoring metric, perturbing the top 25\% of passages successfully changes the predicted answers in about 30\% of examples compared to the 8\% of examples whose answers are successfully changed by perturbing randomly sampled passages. This suggests \textbf{that the model may be ignoring lower-scoring retrieved passages that are less relevant to the query, despite containing the answer entity.}

\paragraph{Confidence Study.}


\begin{table}
\footnotesize
\centering
\begin{tabular}{lcccc}
\toprule
 \%& \multicolumn{2}{c}{NQ-Open}& \multicolumn{2}{c}{TriviaQA} \\
 Pert. & Gen. Prob.& Calib. & Gen. Prob.& Calib.\\
\midrule
25  & 48.15\% & 57.07\% & 62.25\% & 73.72\% \\
50 & 49.67\% & 56.30\% & 70.40\% & 79.90\% \\
75 & 49.86\% & 56.84\% & 73.78\% & 80.59\% \\
100 & 52.22\% & 56.95\% & 74.18\% & 77.61\%\\
\bottomrule
\end{tabular}
\caption{The \% of examples in which model confidence on the correct answer dropped after partial answer substitution in NQ-Open and TriviaQA development set.}\label{tab:confidence-part}
\end{table}

Table~\ref{tab:confidence-part} reports the change in model confidence after performing random entity substitution in the evidence passages. Consistent with the results from~\citet{zhang-etal-2021-knowing}, we find that a separately trained calibrator consistently outperforms the model's inherent confidence score. \textbf{Surprisingly, there is no clear connection between the percentage of perturbed passages and model confidence.}
Ideally, when given a mixed bag of evidence, a model's confidence should decrease to reflect the uncertainty from seeing multiple, conflicting answers. We revisit this in Section~\ref{sec:recalib} where we pilot a calibrator whose confidence drops when presented with conflicting evidence.

\paragraph{Additional Analysis.}
Our confidence study suggests model might not consider all provided passages. To further investigate this, we substitute answers in all passages except top K passages, ranked by the attention score from the reader. Table~\ref{tab:not_topk_attention} presents the results. If you change the answer to all passages except for the top scoring article, the model already outputs the original answer more frequently than the substitute answer.
This again suggests that the model might focus on a handful of most relevant passages and ignore other passages.

\begin{table}
\footnotesize
\begin{center}
\begin{tabular}{rrrrr}
\toprule
& \multicolumn{2}{c}{NQ-Open} & \multicolumn{2}{c}{TriviaQA} \\  
 k &   Original &  Substitute & Original &  Substitute  \\ \midrule
1 & 41.33 & 33.11 & 41.41 & 36.57\\
3 & 69.15 & 12.38  & 66.41 & 19.91\\
5 & 78.50 & 5.97 & 77.11 & 12.71\\
\bottomrule
\end{tabular}

\end{center}
\vspace{-0.5em}
\caption{Substituting all the passages except top k passages (k=[1,3,5]), which are selected based on passage attention scores. On average, 16.7 {(NQ-Open) and 21.5 (TriviaQA) passages out of 100 passages} contained gold answer entity. Yet, with access of up to 3 passages containing the gold answer span, the FiD model can still generate the original answer nearly 70\% of the time.}\label{tab:not_topk_attention}
\vspace{-1em}
\end{table}

In Appendix~\ref{sec:analysis}, we include two further studies. First, we study whether the choice of alternative answer impacts its behaviors. When we provide more realistic alternative answer (either drawn from out-dated corpus or answers to the slightly different interpretation of the question), unsurprisingly, model is \textbf{less} biased to choose the original answer. Second, we study whether model's parametric knowledge is learned during pre-training phase or fine-tuning phase, concluding most of its parametric knowledge is learned during the fine-tuning stage.

\paragraph{Takeaway.} 
 \begin{itemize}[noitemsep,topsep=0pt]
     \item {The models resort to the parametric knowledge to resolve conflicts between different retrieved passages.}
     \item {Model confidence itself cannot be used to identify knowledge conflicts.}
     \item {The model rely on a few most relevant passages, ignoring others.}
 \end{itemize}

\subsection{Adversarial Semantic Perturbation}\label{subsec:semantic}



Semantic perturbation follows earlier work on counterfactual example generation with heuristics~\cite{Ribeiro2020BeyondAB} which perturbs the sentence containing the answer. We simulate four perturbations, and after each perturbation, the model should \textbf{refrain} from returning the original answer. {We aim to test model’s understanding of the passage with such perturbation.} 
\paragraph{Setting.}
We design the four perturbations applicable to question answering: negation, changing to future tense, adding modal verb and text infilling. Examples of each perturbation are in Table~\ref{tab:sub_examples}. To generate these, we run a dependency parser on the sentence containing the gold answer span.\footnote{We use StanfordNLP~\cite{qi2018universal} toolkit.} We then filter examples where the root token of answer sentence is not a verb (about 40\% of sentences, see Appendix~\ref{ssec:appendix-coverage} for full statistics). Finally, we apply simple rules (see Appendix~\ref{ssec:appendix-semantic-details}) to modify the verb. For text infilling, the only difference is that we convert the root token into ``[blank]" and fill in the blank using language modeling \cite{donahue2020enabling}. For passages containing multiple gold answer spans, we apply these perturbations to all sentences as long as their root tokens are verbs.  

\paragraph{Results.}

In Table~\ref{tab:results_semantic_perturb}, we report the exact match to the original answer after applying semantic perturbations. 
{Since our perturbation rules only cover 67-86\% of all sentences containing an answer string, we further subreport our results based on whether there are any remaining unperturbed answer sentences in the evidence. The “partial coverage” subset is the set we created based on the perturbation rules. The “full coverage” subset is created by removing the examples where not all answer sentences have been perturbed.}

Since our perturbation rules only cover 67-86\% of all sentences containing an answer string, we further subreport our results based on whether there are any remaining unperturbed answer sentences in the evidence (partial coverage) or if all answer sentences are perturbed (full coverage). Examples with partial coverage simulate a mixed bag of evidence which may induce the model to return the original answer. In all instances, we expect the exact match to drop significantly after perturbation, as all edits invalidate the original answer; however, we observe that models still return the original answer after perturbation, mirroring what~\citet{Ribeiro2020BeyondAB} finds with extractive models.\footnote{Semantic perturbation details (e.g., statistics of \% of valid examples after perturbation) in the Appendix~\ref{ssec:appendix-coverage}.}


\begin{table}
\small
\centering
\begin{tabular}{lcccc}
\toprule
& \multicolumn{2}{c}{Partial Coverage} & \multicolumn{2}{c}{Full Coverage} \\\midrule
\# passages & 1 & 100 & 1 & 100\\ \midrule

negation  &  82.49 & 86.80 & 74.71 & 71.26\\
modality & 89.90 & 92.48 & 88.77 & 84.05 \\
future & 91.90 & 94.03 & 90.72 & 86.93\\
text-infilling & 88.66 & 93.21 & 86.96 & 84.71 \\
\bottomrule
\end{tabular}
\caption{Exact match score with the original answer after perturbation of each type: models largely disregard the perturbation and outputs the original answer. }\label{tab:results_semantic_perturb}\vspace{-0.5em}
\end{table}




\paragraph{Confidence Study.}
We repeat the calibration study with semantic perturbation. We find that calibration scores remain mostly steady after the perturbation for all four perturbation types, only for 30-40\% of examples we see a decrease in calibration score after the perturbation. The model is particularly less sensitive to temporal perturbation (future). The exact numbers and the ratio of calibration scores before and after the perturbation can be found in the Appendix~\ref{ssec:appendix-confidence}. We observe that model behaves similarly to extractive model~\cite{Ribeiro2020BeyondAB}, returning an answer matching the answer type with high confidence even when the passage no longer supports it.

\begin{table*}\vspace{-0.5em}
\small
\centering
\begin{tabular}{l|rrrr|r}
\toprule 
   \multirow{3}{*}{Perturbation Method} &  Original & Partial  & AmbigQA &  SituatedQA & {Macro} \\  
 & (NQ Dev)& Sub.  &  (Disambiguated Q.) &  (Retrieval 2021) & Average \\ \smallskip
 & ($N$ = 8.7k)  & ($N$ = 2.5k*2) & ($N$ = 448*2) & ($N$ = 55*2) & \\ \midrule
  
{Model Confidence} & \textbf{66.32} & 28.21 & 13.17 & 20.24 & 31.99 \\
{Org. Calibrator ($N_{tr}$ = 80k)} & 62.92 & 12.63 & 4.46 & 8.33 & 22.09 \\ \midrule
{Partial Sub.  ($N_{tr}$ = 36k*2)} & 63.61 & \textbf{98.60} & 13.62 & 28.57 & 51.1  \\
{AmbigQA \;  ($N_{tr}$ = 5k*2)}& 59.84 & 42.15 & \textbf{87.93} & 79.76 & 67.42 \\
{SituatedQA ($N_{tr}$ = 1.5k*2)} & 59.09 & 37.15 & 40.40 & \textbf{92.86} & 57.38 \\
{Part.+Amb.+Sit. ($N_{tr}$ = 42.5k*2)} & 64.55 & 98.22 & 54.46 & 72.62 & \textbf{72.46}\\

\bottomrule
\end{tabular}
\caption{Binary accuracy (\%) of the calibrator trained on different augmented datasets tested on various evaluation sets.  Each column represents evaluation set. Baseline calibrators are in the top row block, and calibrators are trained with augmented training data matching evaluation dataset in the second row block. $N_{tr}$ stands for the training dataset size, {and $N$ denotes the size of each evaluation set.} }\label{tab:recalibration}\vspace{-0.5em}
\end{table*}\section{Re-Calibrating Models Given a Mixed Bag of Evidence}\label{sec:recalib}
When presented with a mixed bag of evidence, systems should inform users of the multiple, conflicting answers. While there are many of approaches for relaying this information to users (e.g., composing a paragraph aggregating answer candidates, or providing set of answers mapped to documents supporting them), a necessary prerequisite to all such systems is the ability to detect when there are conflicting answers in the evidence. Thus, we explore creating systems that can \textbf{detect and abstain} from predicting on instances with conflicting evidence. Questions should only be answered if (1) there is no knowledge conflict in its evidence set and (2) model's predicted answer matches the annotated answer. We report calibrator's binary calibration accuracy following prior work~\cite{Kamath2020SelectiveQA}. We explain four evaluation settings here. 

\noindent \textbf{Original} We use the original NQ development set as is to provide a reference for the performances of calibrators.

In the following settings, we only look at examples where the original FID model correclty answers. Thus, the calibrator should only abstain for knowledge conflict. We construct three different types of knowledge conflict set where calibrator should abstain on \textbf{half} of the examples because of the knowledge conflict. To construct these set, we use the original question, 100 evidence passage set (where model should present its answer), and augment \textbf{one} perturbed example, where 100 evidence passage set is perturbed to have multiple answer candidates to the same question. We discuss three ways to introduce perturbed evidence set, with more than one valid answer candidate now.



\noindent \textbf{Partial Substitution} We use the sets of conflicting evidence passages constructed in Section~\ref{subsec:entity} (randomly sampling 50\% of the retrieved passages to substitute a new entity in).

\noindent \textbf{AmbigQA}
Instead of random new entity, we sample valid alternative answer to the question taken from a different interpretation of the same question from AmbigQA~\cite{min2020ambigqa} dataset. Instead of simply replacing answer in existing passage, we retrieve new passage for each rewritten, disambiguated version of the question.

\noindent \textbf{SituatedQA}
We sample valid alternative answer from either corpus taken from a different time period from SituatedQA~\cite{Zhang2021SituatedQAIE} dataset. We use the same query, but retrieve over two different snapshots of the same corpus (the Wikipedia dump from 2018 and from 2021).

We evenly combine retrieved passages from conflicting answer sets, using the top retrieved passages that contain the respective answer and backing off to the passages with high retrieval scores if not enough passages contain the answer string.

\paragraph{Calibrator} As a baseline, we use same calibration model from our prior study in Section~\ref{subsec:confidence}. We also retrain separator calibrators for each of our three substituted answer types, which are trained by applying the same data augmentation process that was applied to the evaluation set (described above) to training portion of filtered NQ-Open dataset.


\paragraph{Results}
We report the results in Table~\ref{tab:recalibration}. We observe vanilla model confidence outperforms trained calibrator, showing robustness towards out of domain setting. This could be caused by a large gap in accuracy of FiD model for training (80\%) and testing data (52\%). Base calibrators, without data augmentation, struggles substantially, particularly on real world knowledge conflict scenario where it is presented with multiple valid answer candidates (AmbigQA, SituatedQA). Training with data augmentation improves the calibrator's performance; however, this fix does not easily generalize over different methods of collecting conflicting answers and evidence sets. Interestingly, training with more realistic conflicting evidence sets (AmbigQA, Situated QA), while being substantially smaller, generalizes better than simulated conflicting evidence set (Partial Substitution). Training over all types of conflicting evidence sets jointly improves performance over the baseline {calibrators} only modestly compared to the gains from training on data from each method separately. Future work can explore improving calibrator generalization across different knowledge conflict types.

\section{Related Work}
Recent analysis~\cite{Lewis2021QuestionAA, Krishna2021HurdlesTP} pointed the overlap in training and evaluation dataset inflates question answering performances. \citet{Longpre2021EntityBasedKC} showed that the reader model tend to memorize entity answers despite the answer mentions are substituted by another entity. We showed that memorization do occur when the model can only have access to one passage, but can be reduced significantly if the model is trained with multiple passages. Concurrent work~\cite{pan2021contraqa} investigates QA models' robustness to misinformation by providing contradicting contexts. They focus on \textit{generating} conflicting passages, while we focus on understanding how models behave under such settings, including in-depth study of their confidence score. 


Recent works evaluated robustness by minimally perturbing input examples~\cite{kaushik2020learning,gardner-etal-2020-evaluating} to identify models that are invariant under distributional shift. Prior work explored automatically generating such perturbed input (counterfactual data) with heuristics~\cite{Ribeiro2020BeyondAB} or learned models~\cite{wu-etal-2021-polyjuice,Bartolo2020BeatTA,Paranjape2021RetrievalguidedCG}. Recent work~\cite{Du2022SyntheticDA} studies knowledge poisoning for a related task, fact checking. Our perturbation methods are rule-based similar to ~\citet{Ribeiro2020BeyondAB}, but designed specifically for QA task.

\section{Conclusion}
We summarize our findings: 
     \textbf{Do models ground their answers from retrieved document or parametric knowledge? (Section~\ref{sec:span})}
   {Current SoTA models ground their answers mostly from retrieved passages, {when} paired with a high recall retriever (Table~\ref{tab:initial},~\ref{tab:results_entity_perturb}).}

    \textbf{How do models use multiple passages when different passages suggest different answers? (Section~\ref{subsec:entity})} {Models rely on a few, most relevant passages (Table~\ref{tab:not_topk_attention}), and use parametric knowledge to break ties (Figure~\ref{fig:partial}, Table~\ref{tab:nao}).}

    \textbf{How do models behave if some passages are perturbed \textbf{not} to support an answer? (Section~\ref{subsec:semantic})}~{Models largely ignore semantic perturbations and outputs potential answer entity in the retrieved passages (Table~\ref{tab:results_semantic_perturb}).}

    \textbf{How is the model's confidence score affected by knowledge conflicts?} ~{Confidence score is not sensitive to knowledge conflicts (Table~\ref{tab:confidence-part}, Figure~\ref{fig:conf}), and a separately trained calibrator offers some improvements.}

     \textbf{Can we train a model to refrain from returning a single answer when there is conflicting evidence?} {If we train a calibrator on the conflicting evidence set, calibrator can learn to refrain, but does not generalize to different types of conflicting evidence sets (Table~\ref{tab:recalibration}).}
    
     \textbf{What should the model do when there is conflicting evidence?} {We present a partial solution of training a calibrator which learns to abstain from answering when provided conflicting evidence. Future work can explore summarizing and comparing different answers suggested by diverse passages.}

Overall, models' limited ability to aggregate conflicting information among its rich knowledge sources encourage future work in this domain.

\section*{Limitations}
Our study is based on current state-of-the-art model on popular benchmark datasets. For other datasets (e.g., datasets where retrieval quality is substantially worse) or different models~\cite{brown2020language, chowdhery2022palm, rae2021scaling, thoppilan2022lamda} of substantially richer parametric knowledge, our observation that memorization is relatively rare will not hold.

We focus on extractive question answering task, where the answer consists of short entity span. Studying knowledge conflicts in complex question answering tasks where answer is multi-sentence~\cite{fan-etal-2019-eli5} or conditional~\cite{Sun2022ReasoningOL} requires future work.

Lastly, most of our knowledge conflicts study (except the settings where we retrieve passages with AmbigQA and SituatedQA) are simulated, and we leave identifying and evaluating model on real-world knowledge conflicts as future work. 

\section*{Acknowledgements}
We thank members of UT Austin NLP community and Sewon Min for providing feedback in earlier draft of the paper. The work is partially supported by a grant from Open Philanthrophy and Google Research Award.

\bibliography{anthology,custom}
\bibliographystyle{acl_natbib}

\newpage
\appendix
\newpage
\section{Appendix}

\subsection{Calibrator Hyperparameter}\label{subsec:calhyper}

The input to the calibrator is the concatenation of the generation probability and the encoder feature representation averaged across length, and the output is a score indicating the probability of the model correctly predicting the answer. {For each dataset, we reserve 4K examples of the training set for validation,} and trained our calibrator on the remaining data. Hyperparameters are selected based on AUROC on validation set.

We use 100 boosting rounds, subsample ratio of 0.5 and learning rate of 0.5. The same subsample ratio is applied for constructing each tree, for each level and for each split.
\label{sec:appendix}

\begin{table}
\small
\centering
\begin{tabular}{l|c|ccc}
\toprule
\multicolumn{2}{c|}{} & neg. /modal. & negation-& text-\\
\multicolumn{2}{c|}{\#Passages}  &/ fut. & polyjuice & infilling \\\midrule

\multirow{2}{*}{\%Ex.} &  1  & 61.14\% & 57.38\%& 62.43\% \\
  & 100 & 89.55\% & 88.18\% & 89.68\% \\ \midrule
  \multirow{2}{*}{\%Cov.}  & 1 & 85.77\%  & 82.90\% & 86.07\% \\
  & 100 &  66.93\% & 61.12\% & 68.25\% \\ \midrule
  
  \%Ex.       &  1   & 51.87\% & 46.39\%& 53.23\% \\
  (100\% Cov.)& 100. &  15.13\% & 11.89\%  & 16.06\% \\ 
  
  
\bottomrule
\end{tabular}
\caption{Data statistics for different perturbations schemes. The first two rows are the numbers of examples, shown in percentage out of the examples that FiD can answer correctly. The third and fourth rows shows the percentage of gold answer span covered (valid for perturbation) in the chosen examples. The last two rows shows the percentage of valid examples we could get if all the gold answer spans are perturbed.}\label{tab:beyond-data}
\end{table}

\begin{table*}
\small
\begin{center}
\begin{tabular}{r|rr|rr|rr|rr}
\toprule
 & \multicolumn{2}{c|}{Random Entity } & \multicolumn{2}{c|}{AmbigQA Entity} &  \multicolumn{2}{c|}{Random Entity} &  \multicolumn{2}{c}{SituatedQA Entity}  \\

\% & \multicolumn{2}{c|}{(on AmbigQA set)} & \multicolumn{2}{c|}{} &  \multicolumn{2}{c|}{(on SituatedQA set)} &  \multicolumn{2}{c}{}  \\
Perturbed&   Original &  Substitute  & Original &  Substitute  & Original &  Substitute  & Original &  Substitute    \\ \midrule
25  & 74.90 & 6.37   & 75.45 & 9.15  & 76.74 & 4.65  & 80.00 & 7.27  \\
50  & 51.79 & 24.70  & 51.11 & 27.68 & 55.81 & 16.28 & 56.36 & 25.45 \\
75  & 27.88 & 43.03  & 25.22 & 46.21 & 46.51 & 13.95 & 38.18 & 43.64 \\
100 & 2.39  & 65.34  & 5.80  & 63.17 & 4.65  & 39.53 & 14.55 & 58.18 \\
\bottomrule
\end{tabular}
\end{center}
\caption{Entity substitution results on subsets of NQ-Open. We perform random entity substitution on the AmbigQA and SituatedQA sets for fair comparisons between different sources of substitute answers. }\label{tab:appendix-subset}
\end{table*}

\subsection{Model and Training Details}
\label{ssec:appendix-model}
The Fusion-in-Decoder (FiD) model consist of a retriever and a reader module. The retriever~\cite{karpukhin2020dense} is a BERT bi-encoder model, which calculate the similarity between the question $q$ and each of the passages $\{p_i\}$ in the knowledge source and output the most similar ones. The similarity is computed as the dot product of the encoded vectors
$$ E_Q(q)^{T}E_P(p_i) $$
where $E_Q$ is the question encoder and $E_P$ is the passage encoder.

The reader module is a pretrained T5-large~\cite{2020t5}, an encoder-decoder model containing 770M parameters. Each passage is concatenated with the question and truncated to 250 word pieces. For our experiments finetuning FiD, we train the reader module with 1, 20, and 50 evidence passages. To train the reader, we use the AdamW optimizer~\cite{loshchilov2018decoupled} and a learning rate of $5 \cdot 10^{-5}$ with linear warmup of 8000 steps followed by linear decay to zero. The total training steps is 300k, and the final model checkpoint is selected based on exact match score on NQ Open development set. We only use batch size of 1 due to memory constraints. The models take roughly 7 GPU days to train on a Quadro RTX 8000 machine. 

The closed-book question answering (CBQA) model is trained using a T5-large pretrained model, with a batch size of 32, 500k total training steps, and all the other hyperparameters the same as FiD reader models. It roughly take 2 GPU days to train on a Quadro RTX 8000 machine. 

\subsection{Perturbation Coverage}
\label{ssec:appendix-coverage}
As mentioned in Section~\ref{sec:perturbation}, if the root token of the answer sentence is not a verb, then we ignore that sentence, and thus some examples would be excluded. The first row shows the percentage of valid examples after applying the rules mentioned in Section~\ref{sec:perturbation}. We consider it valid example if one of the gold answer span can be perturbed. The corresponding percentage of perturbed gold answer spans is shown in the third row. A small portion of gold answer spans remain unchanged after performing the perturbation. For the second and fourth row it shows the same except the model has access to 100 passages. The percentage of valid examples are much higher since we consider the example valid if one of the gold answer spans in any of the passages can be perturbed. The last two rows show the percentage of examples where all gold answer spans in all the retrieved passages can be perturbed.

\begin{table*}
\small
\begin{center}
\begin{tabular}{r|l|l|l|l|l|l}
\toprule
\%& \multicolumn{2}{c|}{NQ Open}  & \multicolumn{2}{c|}{AmbigQA} & \multicolumn{2}{c}{SituatedQA}\\
perturbed&   Original &  Substitute &   Original &  Substitute &   Original &  Substitute   \\ \midrule
25& 67.35 & 9.51 & 72.16 & 7.21 & 66.67 & 0.00 \\
50 & 45.50 & 27.51 & 40.20 & 34.02 & 33.33 & 33.33\\
75 & 21.85 & 48.84 & 22.68 & 41.23 & 0.00 & 66.67\\
100 & 0.00 & 68.12 & 1.03 & 63.92 & 0.00 & 66.67\\

\bottomrule
\end{tabular}
\end{center}
\caption{Exact match score of substituting different number of passages on NAO sets.}\label{tab:conflicting-nao}
\end{table*}

\subsection{Technical Details on Semantic Perturbations}
\label{ssec:appendix-semantic-details}

For perturbation schemes except text infilling, we first identify the root token's part-of-speech tag. If it is in one of [VB, VBP, VBZ], then we treat it as the present tense, and modify the verb accordingly. (e.g. V $\rightarrow$ "does not V"/"do not V" for negation, V $\rightarrow$ "may V" for modality, V $\rightarrow$ "will V" for future tense) The lemmatized verb forms after "will" and "may" are obtained by the  "WordNetLemmatizer" class in nltk\footnote{\url{https://www.nltk.org/_modules/nltk/stem/wordnet.html}}.  We also identify ["is", "am", "are"] and modify the verbs into their corresponding forms. If the part-of-speech tag is VBD, then it is in past tense and the root token is modified similarly to present tense. Lastly, if the part-of-speech tag is VBN or VBG, then it is present/past participle or gerund. We then identify the be-verbs and/or ["had", "have", "has"], and perform modifications accordingly.

\begin{table}
\small
\begin{center}
\begin{tabular}{r|l|l}
\toprule
\%& \multicolumn{2}{c}{Exact Match}  \\
perturbed&   Original &  Substitute    \\ \midrule
25& 80.00 & 7.27 \\
50 & 60.00 & 25.45\\
75 & 41.82 & 43.64\\
100 & 18.18 & 60.00\\
\bottomrule
\end{tabular}
\end{center}
\caption{Results of substituting different number of passages on SituatedQA. The substitute answer is randomly selected from the SituateQA answer set and is not in the original ansewr set.}\label{tab:conflicting-situated-random}
\end{table}

\subsection{Model Tested on NQ Open Subset}\label{ssec:appendix-nqopen-subset}
Both AmbigQA and SituatedQA annotate subsets of NQ Open. To ensure identical data distribution and isolate the effect of different substitute answers,  we report results of random entity substitution on AmbigQA set and SitutatedQA set respectively. We present the results in Table~\ref{tab:appendix-subset}. For AmbigQA subset, different substitute entity types (random or alternative valid entity) do not seem to affect the results too much. However, the model seems to bias toward the substitute answer more with valid alternative entity substitutions on SituatedQA subset, indicating the parametric knowledge of model do know which answers are more likely to be correct. One possible explanation is that AmbigQA answers do not always take the same form as the original ones (e.g. \textit{76th season} and \textit{1995} in Table~\ref{tab:sub_examples}).

\subsection{Answer Entity Sampling Details}
\label{ssec:appendix-sampling}
When substituting with AmbigQA answers, we consider only the examples with multiple valid answers. For each example, we randomly sample one answer not in the original answer set of NQ as the substitute answer. For substitution with SituatedQA answers, we select the most recent answer as substitute answer. We also include the result of randomly sample an answer from SituatedQA answer set in Table~\ref{tab:conflicting-situated-random}.


\subsection{Full Results on No Answer Overlap Set}\label{ssec:nao-appendix}
Table \ref{tab:conflicting-nao} contain the full results on NAO set for NQ Open, AmbigQA, and SituateQA.

\subsection{Confidence Study Full Results}

\begin{figure}
\centering\vspace{-1.5em}
    \includegraphics[width=7.5cm]{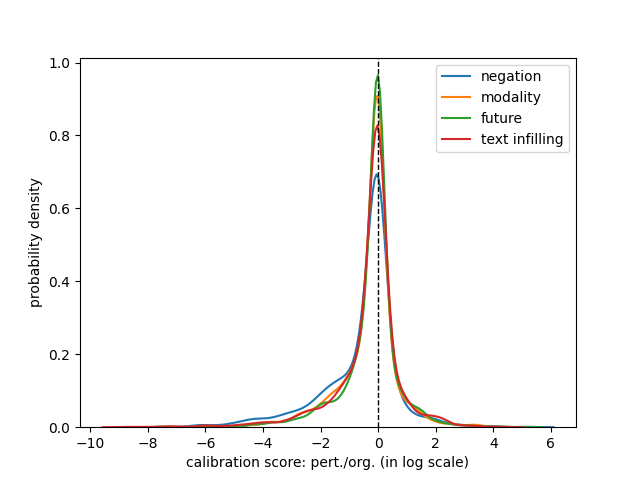}\vspace{-0.5em}
    \caption{The ratio of calibration score after perturbation to that before perturbation, in log scale. The occurrences of examples of different ratio are plotted in terms of probability density (the area under curve is sum to 1). The distributions are bell-shaped, but shift slightly towards negative x-axis.}\vspace{-0.5em}
    \label{fig:conf}
\end{figure}\label{ssec:appendix-confidence}
Table~\ref{tab:confidence} contains the full results for confidence study on adversarial semantic perturbation.
\begin{table}
\small
\centering
\begin{tabular}{lll}
\toprule
%
Change Type & Gen. Prob. & Calibration \\ \midrule
negation  &  65.94\% & 70.28\%   \\
modality & 62.75\% & 66.34\%   \\
future & 58.87\% & 62.92\%   \\
text-infilling & 60.56\% & 64.36\%  \\
\bottomrule
\end{tabular}
\caption{The percentage of examples in which model confidence dropped after perturbation; i.e., the model confidence when predicting the original example is higher than the perturbed example. Model confidence is measured with generation probability/calibration.}\label{tab:confidence}
\end{table}

\subsection{Domain Adaptation Results for Entity Substitution}
\label{ssec:domain-adapt}
{
We would like to study the memorization issue when the model is tested on out-of-domain datasets. Following the setting in Section~\ref{sec:span}, we substitute the answer entity mentions in the retrieved passages with random entities of the same type after the retrieval step. The only difference is that the reader model is trained on a different domain. We evaluate FiD reader model which is trained with NQ-Open on TriviaQA dataset, and vice versa. The results are presented in Table~\ref{tab:domain-adapt-nq-on-tqa} and ~\ref{tab:domain-adapt-tqa-on-nq}. The memorization ratio is still low with high-recall retrievers for both settings, indicating that the model actually relies on the retrieved passages under the distribution shift. }

\section{Further Analysis}
\label{sec:analysis}
We further examine our results, focusing on the quality of substitute answer in entity substitution study and which parametric knowledge (pre-training vs. fine-tuning) was used.
\paragraph{Improving Substitute Entities}
Prior work~\cite{Longpre2021EntityBasedKC} substitutes answer entity with another entity with same coarse entity type. This makes substitute entities sometimes unreasonable, despite better than randomly sampling entities without type constraint. For example, ``Heartbreak Hotel" was substituted as an answer to the following question ``who did the lions play on thanksgiving last year''.

We make perturbation more realistic by substituting with alternative answer from two datasets, AmbigQA~\cite{min2020ambigqa} and SituatedQA~\cite{Zhang2021SituatedQAIE}, which augmented existing NQ open dataset. Both datasets annotated {valid} alternative answers for different interpretation of the same question (AmbigQA) and answers belonging to different temporal contexts (SituatedQA) for NQ-Open dataset. We sample these additional answers as a new answer to inject (details in Appendix~\ref{ssec:appendix-sampling}).


\begin{table}
\centering
\small
\begin{tabular}{lllrrrr}
\toprule
Model &\# Pass. & \% & Ans.  & \multicolumn{2}{c}{Exact Match} & \multirow{2}{*}{$M_R$}  \\ 
&train / inf. & ex.& R &  Orig. &  Sub.  & \\ \midrule
FiD & 1 / 1 & 28.4 & 48.5 & 12.5 & 42.0 & 22.9 \\ 
FiD & 5 / 1 & 28.7 & 72.9 & 5.1 & 54.0 & 8.7 \\ 
FiD & 5 / 5 & 33.7 & 72.9 & 5.4 & 52.8 & 9.2 \\ 
FiD & 20 / 1 & 27.9 & 83.1 & 3.8 & 62.6 & 5.8 \\ 
FiD & 20 / 20 & 35.0 & 83.1 & 4.3 & 60.4 & 6.7 \\ 
FiD & 50 / 1 & 29.2 & 86.8 & 3.8 & 63.9 & 5.6 \\ 
FiD & 50 / 50 & 37.6 & 86.8 & 4.7 & 62.4 & 7.0 \\ 
FiD & 100 / 1 & 28.7 & 88.7 & 4.6 & 61.9 & 6.9 \\ 
FiD & 100 / 100 & 37.1 & 88.7 & 5.5 & 58.0 & 8.6 \\ 
\bottomrule
\end{tabular}
\caption{Exact Match / Memorization Ratio for FiD
model trained on NQ-Open with different amount of passages and evaluated on TriviaQA. The memorization is still low for domain adapted models, when provided with multiple retrieved passages.}
\label{tab:domain-adapt-nq-on-tqa}
\vspace{-0.5em}
\end{table}

\begin{table}
\centering
\small
\begin{tabular}{lllrrrr}
\toprule
Model &\# Pass. & \% & Ans.  & \multicolumn{2}{c}{Exact Match} & \multirow{2}{*}{$M_R$}  \\ 
&train / inf. & ex.& R &  Orig. &  Sub.  & \\ \midrule
FiD & 1 / 1 & 17.0 & 67.1 & 12.8 & 51.1 & 20.0 \\ 
FiD & 5 / 1 & 17.1 & 81.7 & 6.7 & 63.6 & 9.5 \\ 
FiD & 5 / 5 & 21.1 & 81.7 & 5.6 & 57.4 & 9.0 \\ 
FiD & 20 / 1 & 16.9 & 85.7 & 5.4 & 65.1 & 7.6 \\ 
FiD & 20 / 20 & 22.1 & 85.7 & 3.9 & 56.1 & 6.6 \\ 
FiD & 50 / 1 & 17.0 & 87.2 & 3.9 & 69.7 & 5.2 \\ 
FiD & 50 / 50 & 22.4 & 87.2 & 3.8 & 61.0 & 5.9 \\ 
FiD & 100 / 1 & 16.3 & 87.9 & 5.5 & 65.6 & 7.7 \\ 
FiD & 100 / 100 & 22.5 & 87.9 & 3.5 & 59.2 & 5.6 \\
\bottomrule
\end{tabular}
\caption{Exact Match / Memorization Ratio for FiD
model trained on TriviaQA with different amount of passages and evaluated on NQ-Open. }
\label{tab:domain-adapt-tqa-on-nq}
\vspace{-0.5em}
\end{table}

\begin{table}
\small
\begin{center}
\begin{tabular}{r|rr|rr}
\toprule
Entity source &  \multicolumn{2}{c|}{AmbigQA  (N=448)} &  \multicolumn{2}{c}{SituatedQA  (N=55)}  \\
 \% per.&   Ori. &  Sub. & Ori. &  Sub.    \\ \midrule
25  & 74.11 & 10.40  & 77.45 & 8.36  \\
50  & 50.71 & 26.65 & 56.73 & 25.09 \\
75  &  25.40 & 47.05 & 33.09 & 44.73 \\
100 &  5.80  & 63.17 & 14.55 & 58.18 \\
\bottomrule
\end{tabular}
\end{center}
\caption{Results of substituting different proportion of 100-retrieved passages on NQ-Open where entities are derived from AmbigQA and SituatedQA dataset. The number next to the entity refers to the number of examples in this evaluation set after filtering. }\label{tab:newentityperturb}
\end{table}

\begin{table}\begin{center}
\small
\begin{tabular}{rlrrr}
\toprule
\% per. & Dataset &   NAO&  AO  &  AO\%  \\ \midrule
50\% &TQA (Random Entity)  &  75.25 & 81.40 & 86.66\\
50\% &NQ (Random Entity)  & 61.83  & 70.92 & 85.93\\

50\% & w/ AmbigQA Entity  & 55.08 & 66.40 & 78.35\\
50\% & w/ SituatedQA Entity  & 50.00  & 71.36 & 94.55\\ \midrule

100\% &TQA (Random Entity) & 2.75  & 9.37 & 86.66\\
100\% &NQ (Random Entity) & 0.45  & 4.14 & 85.93\\
100\% & w/ AmbigQA Entity & 1.59 & 10.16 & 78.35\\
100\% &w/ SituatedQA Entity & 0.00 & 21.05 & 94.55\\ 
\bottomrule
\end{tabular}
\end{center}\vspace{-0.6em}
\caption{Memorization ratio ($M_R$ of substituting different number of passages on NQ-Open No Answer Overlap (NAO) / Answer overlap (AO) set {of NQ-Open and TriviaQA}. {AO\% signifies the percentage of examples that belong to AO set for each subset.}}\label{tab:nao}\vspace{-0.5em}
\end{table}



Table~\ref{tab:newentityperturb} presents perturbation results with valid entities sourced from AmbigQA and SituatedQA. We identify a surprising trend -- that model outputs original answers more frequently when substituted with better alternatives. This contradicts our intuition as model should be less hesitant to choose new substitute answer as they are also valid answer to the question, for different contexts. We further investigate this issue below. 



\paragraph{Does parametric knowledge come from pre-training or fine-tuning?}
\label{subsec:param}
Some memorization (2--15\%) remains even after all the evidence documents are perturbed, and model is biased toward the original answer under partial substitution. We aim to identify whether it comes from pretraining or fine-tuning of the reader model by using the evaluation data splits from prior work~\cite{Lewis2021QuestionAA}: questions where answers were seen (Answer Overlap (AO)) and questions where answers were unseen (No Answer Overlap (NAO)). If memorization ratio is higher on AO set compared to NAO set, we can hypothesize that memorization mostly happens during fine-tuning compared to pre-training.\footnote{Earlier study~\cite{Longpre2021EntityBasedKC} in a single document setting also reports memorization is more severe in AO set.}  

Table~\ref{tab:nao} presents results for 50\% and 100\% substitution setting.\footnote{See Appendix~\ref{ssec:nao-appendix} for 25\% and 75\% substitution setting.} This study shed lights on mysterious trend: there were more examples with answer overlap in AmbigQA/SituatedQA subset. If we perturb all the evidence documents, the model exhibit little to no memorization on NAO portion. We can thus infer that memorization effect comes almost exclusively from fine-tuning. When accounting for different proportion of answer overlap examples in the subsets, memorization ratio is \textbf{lower} in AmbigQA/SituatedQA NAO set. This suggests that model uses parametric knowledge -- which answer candidate is more reasonable -- in a subtle way, even when behaving as a copying model.

\end{document}